\documentclass[runningheads]{llncs}
\usepackage{graphicx}
\usepackage{amsmath,amssymb} 
\usepackage{color}
\usepackage{csquotes}
\usepackage{multirow}
\usepackage{array}

\usepackage[normalem]{ulem}

\newcolumntype{P}[1]{>{\centering\arraybackslash}p{#1}}
\newcolumntype{M}[1]{>{\centering\arraybackslash}m{#1}}

\newcommand\eg{\emph{e.g.}} 
\newcommand\ie{\emph{i.e.}}

\begin{document}
\pagestyle{headings}
\mainmatter

\def\ACCV20SubNumber{***}  

\title{3D Guided Weakly Supervised Semantic Segmentation} 
\titlerunning{3D Guided Weakly Supervised Semantic Segmentation}
%

\author{Weixuan Sun$^1$ $^2$, Jing Zhang$^1$ $^2$, Nick Barnes$ ^1$}
\institute{$^1$ Australian National University, Australia; $^2$CSIRO Data61, Australia}

\authorrunning{W Sun, J Zhang, N Barnes}

\maketitle

\begin{abstract}
Pixel-wise clean annotation is necessary for fully-supervised semantic segmentation, which is laborious and expensive to obtain.
In this paper, we propose a weakly supervised 2D semantic segmentation model by incorporating sparse bounding box labels with available 3D information, which is much easier to obtain with advanced sensors.
We introduce a 2D-3D inference module to generate accurate pixel-wise segment proposal masks.
Guided by 3D information, we first generate a point cloud of objects and
calculate a per class objectness probability score for each point using projected bounding-boxes.
Then we project the point cloud with objectness probabilities back to the 2D images followed by a refinement step to obtain segment proposals, which are treated as pseudo labels to train a semantic segmentation network.
Our method works in a recursive manner to gradually refine the above-mentioned segment proposals. 
We conducted extensive experimental results on the 2D-3D-S dataset where 
we manually labeled a subset of images with bounding boxes.
We show that the proposed method can generate accurate segment proposals when bounding box labels are available on only a small subset of training images. 
Performance comparison with recent state-of-the-art methods further illustrates the effectiveness of our method. 

\end{abstract}

\section{Introduction}
\label{introduction}

Recent work on 2D image semantic segmentation has achieved great progress via adopting deep fully convolutional neural networks (FCN) \cite{long2015fully}. 
The success of these models \cite{zhao2017pyramid,chen2017deeplab,chen2014semantic,zhao2018psanet,chen2017rethinking} arises from large training datasets with pixel-wise labels, 
which are laborious and expensive to obtain. For example, the cost of pixel-wise segmentation labeling is 15 times larger than bounding box labeling and 60 times larger than image-level labeling \cite{lin2014microsoft}.

\begin{figure}
\setlength{\abovecaptionskip}{1mm}  
\centering
\includegraphics[width=0.95\linewidth]{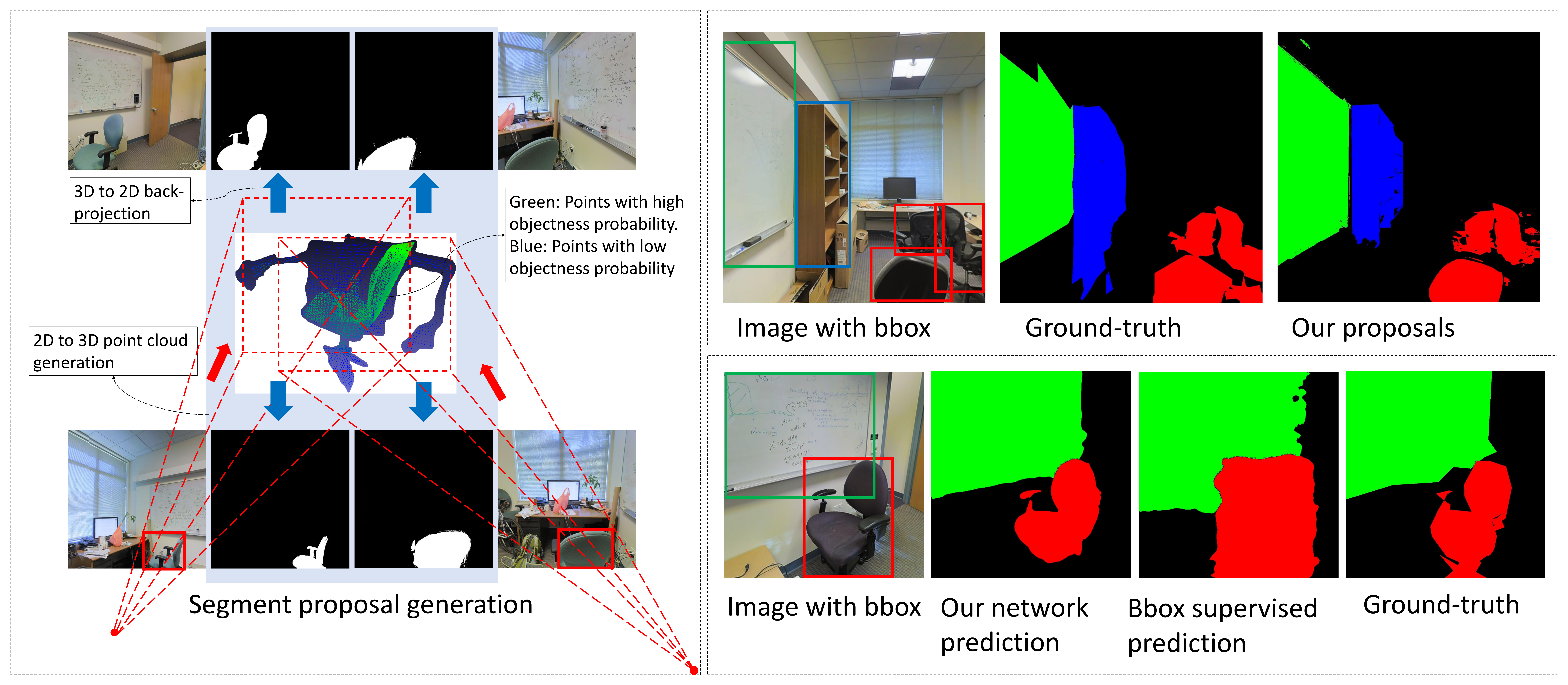}
\caption{Left: Our segment proposal generation pipeline. Top right: Example images of our segment proposals compared with ground-truth. Bottom right : Network prediction example supervised with our segment proposals compared with prediction supervised with bounding box masks and ground-truth segmentation map. The sample images are from the 2D-3D-S dataset \cite{2017arXiv170201105A}.
}
\label{fig:example}
\end{figure}

Unlabeled or weakly-labeled data can be collected in a much faster and cheaper manner, which makes weakly supervised semantic segmentation a promising direction to develop. Multiple types of weak labels have been studied, including image-level labels \cite{huang2018weakly,wei2018revisiting,ahn2018learning,fan2018cian}, points \cite{bearman2016s},
scribbles \cite{vernaza2017learning,lin2016scribblesup,tang2018regularized}, and bounding boxes \cite{dai2015boxsup,papandreou2015weakly,khoreva2017simple,li2018weakly,song2019box}.
Bounding box annotation offers a simple yet intuitive direction,
that is relatively inexpensive, while still offering rich semantic regional information.
Current bounding box based methods \cite{dai2015boxsup,papandreou2015weakly,khoreva2017simple,li2018weakly,song2019box} usually adopt non-learning methods like Conditional Random Fields (CRF) 
\cite{krahenbuhl2011efficient}, GrabCut \cite{rother2004grabcut} or Multiscale Combinatorial Grouping (MCG) 
\cite{pont2016multiscale} to obtain segment proposals, which are then treated as pseudo labels to train semantic segmentation models.

It has been argued that 3D information plays an important role in scene understanding, but most previous semantic segmentation approaches operate only on individual 2D images. 
With more recent data collection technology and sensors, collection of large scale 3D datasets is no longer a cumbersome process. 
Not only 2D RGB information but also accurate corresponding 3D information like depth maps, camera trajectories, and point clouds are collected.
Especially for indoor scene understanding, datasets like 2D-3D-S \cite{2017arXiv170201105A}, SUN3D \cite{xiao2013sun3d}, ScanNet \cite{dai2017scannet} are available. For outdoor autonomous driving there are datasets like Kitti \cite{Geiger2013IJRR}, ApolloScape \cite{huang2018apolloscape} and the Waymo open dataset \cite{sun2020scalability}. 
With the above-mentioned widely available data,
it's natural to raise a question: \enquote{\textit{Can we retain comparably good performance while only labeling a few images by using box-level weak supervision together with 3D information?}}

In this paper, we investigate the task of combining bounding box labels with 3D information for weakly supervised semantic segmentation, aiming at reducing annotation cost
by leveraging available 3D information.
We investigate this by using the Stanford 2D-3D-Semantics dataset (2D-3D-S) \cite{2017arXiv170201105A}.
We propose a novel 3D guided weakly supervised semantic segmentation approach, where a small number of images are labeled with bounding boxes and these images have their corresponding 3D data. 
Our approach can extract segment proposals from bounding boxes on labeled images and creates new segment proposals on unlabeled images of the same object instance.
These proposals are then used to train a semantic segmentation network. 
Further, our approach works in a recursive manner to gradually refine the above-mentioned segment proposals,
leading to improved segmentation results.

The proposed pipeline
(2D-3D inference module)
is shown in the left of Figure \ref{fig:example}, where
we use a chair as an example.
First, we label the chair from two camera viewpoints and extrude bounding boxes from 2D to 3D space to generate a point cloud of the chair. Then we perform 3D inference to compute an objectness probability for each 3D point, representing the possibility of each point belonging to an object.
The objectness probability is computed based on detection frequency across bounding boxes to enhance correct points and suppress noise. As displayed in the left image of Figure \ref{fig:example}, green and blue points have high and low objectness probability respectively.
We then project from the point cloud with objectness probabilities back to the 2D images to obtain
objectness probability masks. Besides the labeled images, we can also back-project the point cloud to new images without labels.
During projection, we propose a novel strategy by using depth maps to deal with occlusion.
Finally, we refine the objectness probability masks to obtain our final segment proposals. 
We evaluate our method on the 2D-3D-S dataset \cite{2017arXiv170201105A}, and experimental results show that our method
considerably outperforms the competing methods using bounding box labels. 
We summarize our contributions as follows:
\begin{itemize}
  \item We propose a 3D-guided, 2D weakly supervised semantic segmentation method. Our method leverages information that is widely available from 3D sensors without hand annotation to yield improved semantic segmentation with lower annotation cost. 
  \item 
  We present a novel 2D-3D probabilistic inference algorithm, which
  combines bounding-box labels and 3D information to simultaneously infer pixel-wise segment proposals for the labeled bounding boxes and unlabeled images.
  \item
  Our 3D weakly supervised semantic segmentation model 
  learns an initial classifier from segment proposals, then uses the 2D-3D inference to transductively generate new segment proposals,
  resulting in further improvements to 
  network performance in an iterative learning manner.
  \item 
  To the best of our knowledge, it is the first work that uses 3D information to assist weakly supervised semantic segmentation.
To evaluate our method we augment the 2D-3D-S dataset \cite{2017arXiv170201105A} with bounding box labels. We demonstrate that our method outperforms competing methods with fewer labeled images.
\end{itemize}

\section{Related Work}
\label{related work}
We briefly introduce existing fully and weakly supervised semantic segmentation models, and 3D information guided models.\\*
\textbf{Fully Supervised Semantic Segmentation}
A series of work has been done based on FCN \cite{long2015fully} for fully supervised semantic segmentation. \cite{long2015fully,ronneberger2015u,badrinarayanan2017segnet} use skip architectures to connect earlier convolutional layers with deconvolutional layers,
to reconstruct fine-grained segmentation shapes. 
The DeepLab series \cite{chen2014semantic,chen2017deeplab,chen2017rethinking} use dilated (atrous) convolution in the encoder, which increases the receptive field to consider more spatial information.
In addition, many methods \cite{liu2015parsenet,zhao2017pyramid,chen2017deeplab,zhao2018psanet,yuan2018ocnet,zhang2019co,zhou2019context,he2019adaptive,fu2019dual} improve semantic segmentation performance by adopting context information. 
\cite{zhao2017pyramid} proposes pyramid pooling to obtain both global and local context information. 
An adaptive pyramid context network is proposed in \cite{he2019adaptive} to estimate adaptive context vectors for each local position.\\*
\textbf{Weakly Supervised Semantic Segmentation}
A large number of weakly supervised semantic segmentation methods have been proposed to achieve a trade-off between labeling efficiency and network accuracy. They usually take low-cost annotation as a supervision signal, including image-level labels \cite{huang2018weakly,wei2018revisiting,ahn2018learning,fan2018cian,papandreou2015weakly}, scribbles \cite{vernaza2017learning,lin2016scribblesup,tang2018regularized}, points \cite{bearman2016s}, and bounding boxes \cite{dai2015boxsup,papandreou2015weakly,khoreva2017simple,li2018weakly,song2019box}. 
Current bounding-box based methods
extract object segment proposals from bounding boxes, which are then used as a network supervision signal.
WSSL \cite{papandreou2015weakly} proposes an expectation-maximization algorithm with a bias to enable refined estimated segmentation maps throughout training. 
BoxSup \cite{dai2015boxsup} proposes a recursive training procedure, which uses generated proposals as supervision in every iteration. \cite{khoreva2017simple} generates segment proposals by incorporating GrabCut \cite{rother2004grabcut} and MCG \cite{pont2016multiscale}.
Most recently, \cite{song2019box}
generate segment proposals with dense CRF \cite{krahenbuhl2011efficient}, and proposes box-driven class-wise masking with a filling rate guided adaptive loss in the training procedure.
However, none of the above methods adopt 3D information.\\*
\textbf{3D Information Guided Semantic Segmentation}
Different from classic 2D RGB semantic segmentation, some work adopts 3D information such as depth maps and point clouds. 
\cite{ren2012rgb,gupta2013perceptual,silberman2012indoor} design handcrafted features tailored for RGB with depth information, extracted features are fed into further models.
\cite{eigen2015predicting,long2015fully} take the depth map as an extra input channel with the RGB images. 
More recently, \cite{gupta2014learning,qi20173d,park2017rdfnet,wang2018depth} encode depth maps into three-dimensional HHA (horizontal disparity, height above ground, and angle with gravity). 
\cite{hou20193d} employs 3D convolutions to extract 3D geometry and 3D colour features from point clouds and project them back to 2D images for segmentation. 
Meanwhile, 3D data is becoming increasingly available from advanced 3D sensors, \eg, \cite{2017arXiv170201105A,xiao2013sun3d,dai2017scannet,Geiger2013IJRR,huang2018apolloscape,sun2020scalability,vechersky2018colourising,chen2019scanrefer,Matterport3D} without requiring human intervention. Which offers opportunity to reduce labeling cost by exploiting automatically obtained 3D information.
We propose to bring in 3D information to assist proposal generation from bounding boxes and propose a 3D guided weakly supervised semantic segmentation network.
As far as we know, this is the first work that combines box-level labels and 3D information for weakly supervised semantic segmentation.

\section{Proposed Approach}
\label{methods}
\begin{figure*}
\setlength{\abovecaptionskip}{1mm}  
   \begin{center}
   {\includegraphics[width=0.9\linewidth]{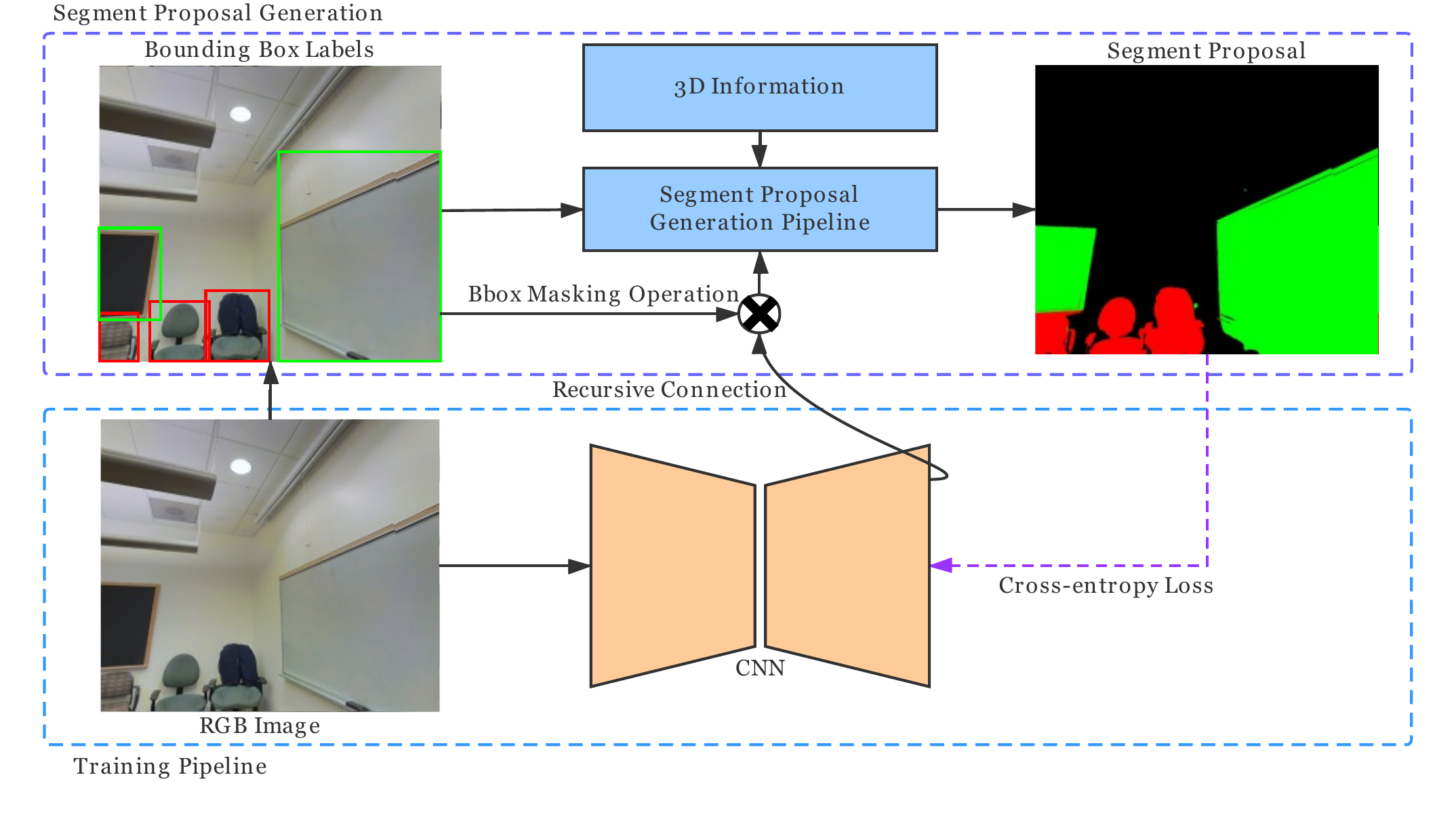}} 
   \end{center}
\caption{Pipeline of the proposed method. We first hand-label a subset of all RGB images with bounding box, which is then
fed into a 3D semantic projection module to generate segment proposals assisted by 3D information. The generated segment proposals are used as supervision signal for our semantic segmentation network. 
Meanwhile, the network predictions can also be fed back into the 3D semantic projection module after the bounding box masking operation in a recursive manner. The details of segment proposal generation pipeline are shown in Figure \ref{fig:3d projection module}.
}
   \label{fig:method_pipeline}
\end{figure*}

We propose a 3D information guided bounding-box based weakly supervised semantic segmentation network.
Specifically, our method consists of two modules: 1) a segment proposal generation module that adopts the 3D information and bounding box labels, and 2) a semantic segmentation network, which takes 2D images as training data. 
First, we feed images with box-level labels and their corresponding 3D information into our segment proposal generation framework, which extracts pixel-wise segment proposals from labeled bounding boxes and generates proposals on new images without labels.
Then, the segment proposals are fed into the training pipeline as a supervision signal to train a semantic segmentation network. 
The network predictions are then fed back into the segment proposal generation module, which generates new segment proposals in a recursive manner.
The entire procedure is shown in Figure \ref{fig:method_pipeline}.

Considering a collection of images $X= \{X_1,..., X_i,...,X_N\}$,
each image has corresponding 3D information, \ie, camera parameters $M_i = (R_i, \Tilde{C_i}, f_i)$ and a depth map $X_{depth}$, where $R_i$ is camera rotation matrix, $\Tilde{C_i}$ is camera position and $f_i$ is focal length. 
In our method, we assume that only a subset of all images are labelled with bounding boxes,
where $X_b \subset X$. 
During labeling, we label images from different camera viewpoints. 
Then we feed the labels into our segment proposal generation framework with their corresponding camera parameters and depth maps. 
We present our segment proposal generation module as a function $S = F(X, X_b, M_i, X_{depth})$, where $S$ is the collection of all segment proposals $S = \{S_1,...,S_i, ..., S_N\}$.
Then the segment proposals $S$ are used supervise the semantic segmentation network: 
$L = L_s(X_p, S)$,
where $L_s$ denotes segmentation loss and  $Y_p$ denotes network prediction. 
In addition, in the recursive process, images labeled with bounding boxes $X_b$ can be replaced with the network predictions $X_p$ from the previous iteration, which is displayed as a recursive connection in Figure \ref{fig:method_pipeline}. The recursive process is introduced in detail in Sec. \ref{recursive procedure}.

\subsection{Segment Proposal Generation:}
\label{segment proposal generation}
\begin{figure}[t!]
\setlength{\abovecaptionskip}{1mm}  
   \begin{center}
   {\includegraphics[width=0.90\linewidth]{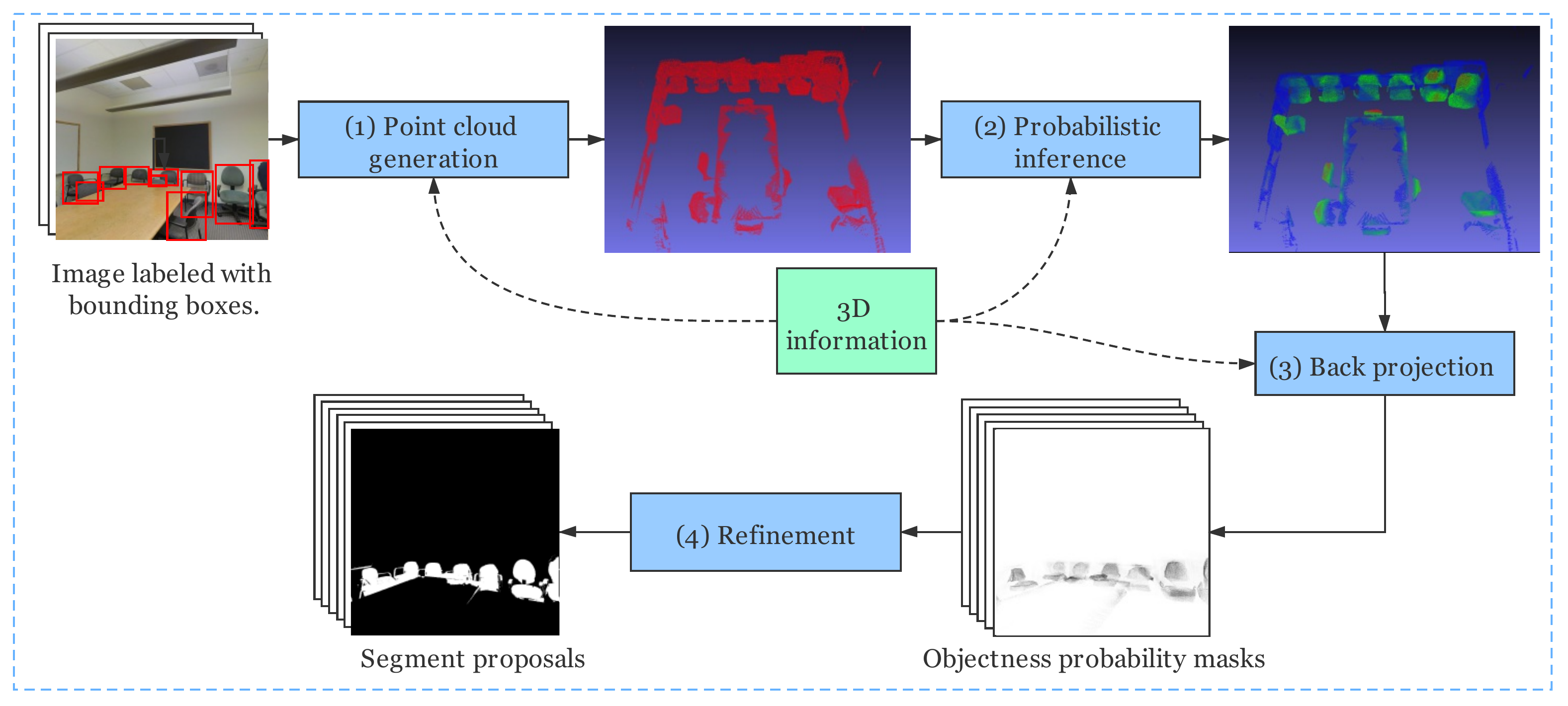}} 
   \end{center}
\caption{Our segment proposal generation pipeline. Taking the class of chair as an example, we first project the point cloud for chairs in this room from bounding box labels. Then, we calculate each point's objectness score. Finally, we back-project the point cloud with probabilities to 2D images followed by refinement to get final segment proposals. 
Note that we can obtain segment proposals on extra images, which means we only need to label a small portion of images. }
   \label{fig:3d projection module}
\end{figure}

Given a set of images $X$ with corresponding 3D information, and annotated bounding boxes for a subset  $X_b$ of the images,
our framework can learn semantic masks $S$ (segment proposals) for all images of a room.
As shown in Figure \ref{fig:3d projection module}, our approach uses a sequence consisting of four components: 
(1) bounding boxes to 3D projection to generate point clouds; (2) 3D probabilistic inference to accentuate correct points and diminish noise; (3) point-wise 3D to 2D projection to generate scattered objectness probability masks; and (4) mask refinement to get final segment proposals. We introduce each component in the following.

\textbf{Point Cloud Generation from Bounding Boxes}
This process aggregates label information from different camera viewpoints into a globally consistent 3D space\footnote{3D information
provided by the 2D-3D-S dataset \cite{2017arXiv170201105A} is determinate,
and SLAM reconstruction is mature, so high quality 3D information is assumed. Our method is not based on SLAM and every point is projected independently so we don't need to handle accumulated errors.}.
Concretely, given an image $X_i$, 
pixels inside bounding boxes are represented as $x^i$.
For one single pixel at position $j$ inside the bounding boxes $x^i_j \in x^i$, we project it into 3D by:

\begin{equation}
P_j^i =[R_i \mid -R_i\Tilde{C_i} ]^{-1} K_i^{-1} * x_j^i * d_j^i, \quad\textrm{where}\quad    K_i^{-1} = \begin{bmatrix}
      \frac{1}{f^i} &   & -\frac{p_x^i}{f^i}   \\
      & \frac{1}{f^i}  & -\frac{p_y^i}{f^i}    \\
      &   &  1
\end{bmatrix}
.
\end{equation}

We follow the finite camera projection model \cite{hartley2003multiple} to project the pixel $x^i_j$ into 3D space, $K_i^{-1}$ denotes inverse camera matrix which projects pixels from the 2D image to camera coordinate, $f^i$ is focal length and $[p_x^i, p_y^i]$ is principal point. $[R_i \mid -R_i\Tilde{C_i} ]^{-1}$ transforms points from camera coordinates to world coordinates, where $R_i$ is a $3\times3$
rotation matrix representing the orientation of the camera coordinates, and $\Tilde{C_i}$ denotes the position of the camera in world coordinates. $d_j^i$ denotes depth information at position $j$ of image $i$.
Then we combine the projected 3D point clouds $P^i$ from different camera views together in world coordinate to obtain a class-specific point cloud.
We perform projection for each labeled bounding box,
label information from different directions and classes is fused into a single 3D point cloud. 
By adopting depth maps, only points nearest to the camera are projected, which ensure accurate object shapes and occluded points are ignored.
The class of every point is decided by the class of the projecting bounding box, so the point clouds are semantically classified.
As shown in Figure \ref{fig:3d projection module}, the red point cloud displays all chairs in the environment. 

\textbf{Point Cloud Probabilistic Inference}
Bounding boxes consist of object and background regions.
When we project pixels in the bounding boxes into 3D space, the background regions are also projected as background noise. As shown by the red point clouds of Figure \ref{fig:3d projection module}, wall and table are also projected and wrongly categorized as chair. 
In order to distinguish points that belong to objects or background,
we take advantage of our multiple views and 3D information.

We propose a novel method to sum objectness confidence across multiple views, which emphasizes the correct points and weakens the irrelevant points (background noise). 
We quantitatively present every point's correctness with a score called the objectness score. 
Inspired by \cite{qi2018frustum}, given that the projection matrix and depth map are known, we can get a 3D bounding frustum from a 2D bounding box, which is the \enquote{visible space} of that bounding box. 
All points inside this 3D bounding frustum can be projected back into the bounding box on 2D image. 
Thus, we make an assumption that, for some camera viewpoints, if a point projected from one camera viewpoint can also be \enquote{seen} by other camera viewpoints, \ie, the point can be projected back into the bounding boxes of other images, the objectness score of the point is higher. 
In this case, with multiple bounding boxes on a single object, the 3D points with a higher objectness score are more likely to belong to this object than to the background.
Specifically, for a 3D point $P_j$ of class $c$, its objectness score $O_j$ is defined as:

\begin{equation}
\label{obj_score}
O_j(P_j \mid B_c) = \sum_{k=1}^{K} F_o(P_j, b_k),
\end{equation}
where $B_c$ denotes bounding boxes of class $c$ in this room and $F_o(P_j, b_k)$ is: 
\begin{equation}
F_o(P_j, b_k) = \begin{cases}
    1   &  \text{if back-projected $P_j$ is inside $b_k$} \\
    0   &  \text{otherwise}\\
 \end{cases} 
 .
\end{equation}

$O_j$ in Eq. \ref{obj_score} indicates the frequency that the point $P_j$ is projected back to bounding boxes across all the images from different viewpoints. We implement the above method for each class independently, then normalize objectness scores for each class to obtain an objectness probability $p(O_j)$:
$p(O_j) = \frac{O_j}{max(O^c)}$, where $O^c$ denotes the collection of objectness scores of all points that belong to class $c$.
Points with higher objectness scores have more confidence to belong to objects.
By doing so, label information from different camera viewpoints are aggregated on points to compose their objectness probability, and reveal objects' shapes. As shown in the top right figure of Figure \ref{fig:3d projection module}, after probabilistic inference, the chairs stand out from background noise, while wall and tables are suppressed.

\textbf{Segment Proposal Generation by Point Cloud Back-projection}
In this stage, we apply 3D to 2D back-projection to generate prototype objectness score masks.
Specifically, given a single point  $P_j$ in the point cloud, we project it to a image at camera viewpoint $i$ \cite{hartley2003multiple}:

\begin{equation}
x_j^i = K_i[R_i \mid -R_i\Tilde{C_i} ]P_j, \quad \textrm{where} \quad   K_i = \begin{bmatrix}
    f_i &   & p_x^i     \\
      & f_i & p_y^i   \\
      &   & 1    
\end{bmatrix}
.
\end{equation}

\begin{figure}
\setlength{\abovecaptionskip}{1mm}  
   \begin{center}
   {\includegraphics[width=0.9\linewidth]{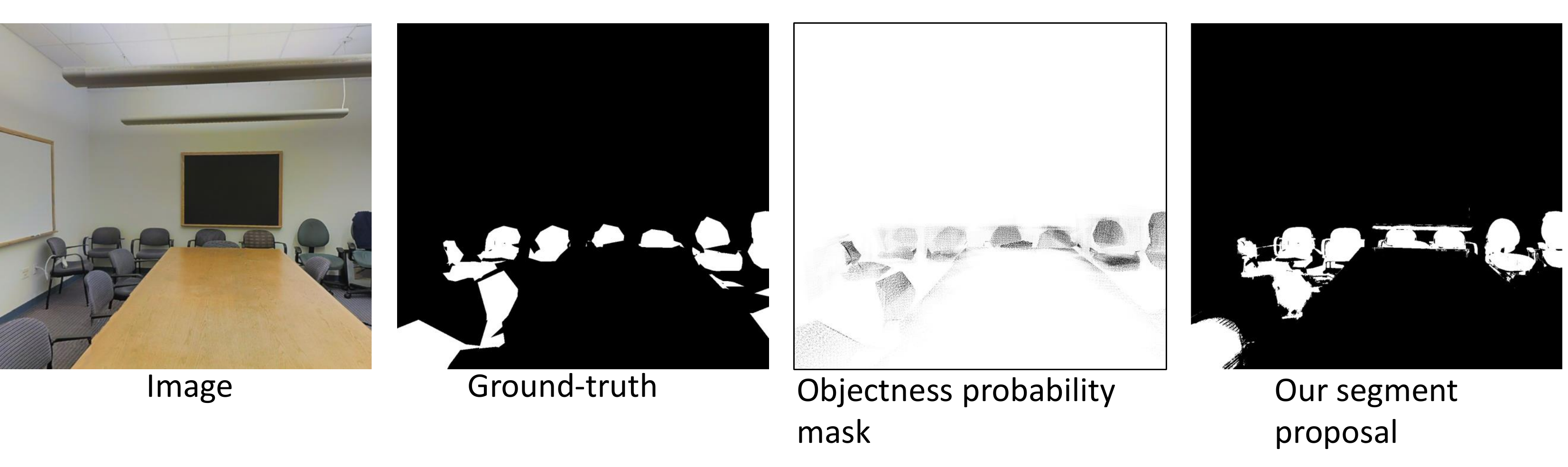}} 
   \end{center}
\caption{Examples of the segment proposals for the class of chair, where we reverse the grayscale of objectness probability masks for better visualization.
}
   \label{fig: proposal refine}
\end{figure}

We project all points with their objectness probabilities and semantic labels onto 2D images.
All 3D points are in the same world coordinate system and can be back-projected to 2D images at any camera viewpoints, no matter whether the images are labeled or not. Therefore, we only need to label a small portion of the images, which can
significantly alleviate annotation cost.
In addition, since the point cloud is sparse in 3D space, we get 2D masks with scattered points and every point represents normalized objectness probability, which are named objectness probability masks, as shown in the third column of Figure \ref{fig: proposal refine}.

\begin{figure}
\setlength{\abovecaptionskip}{1mm}  
   \begin{center}
   {\includegraphics[width=0.8\linewidth]{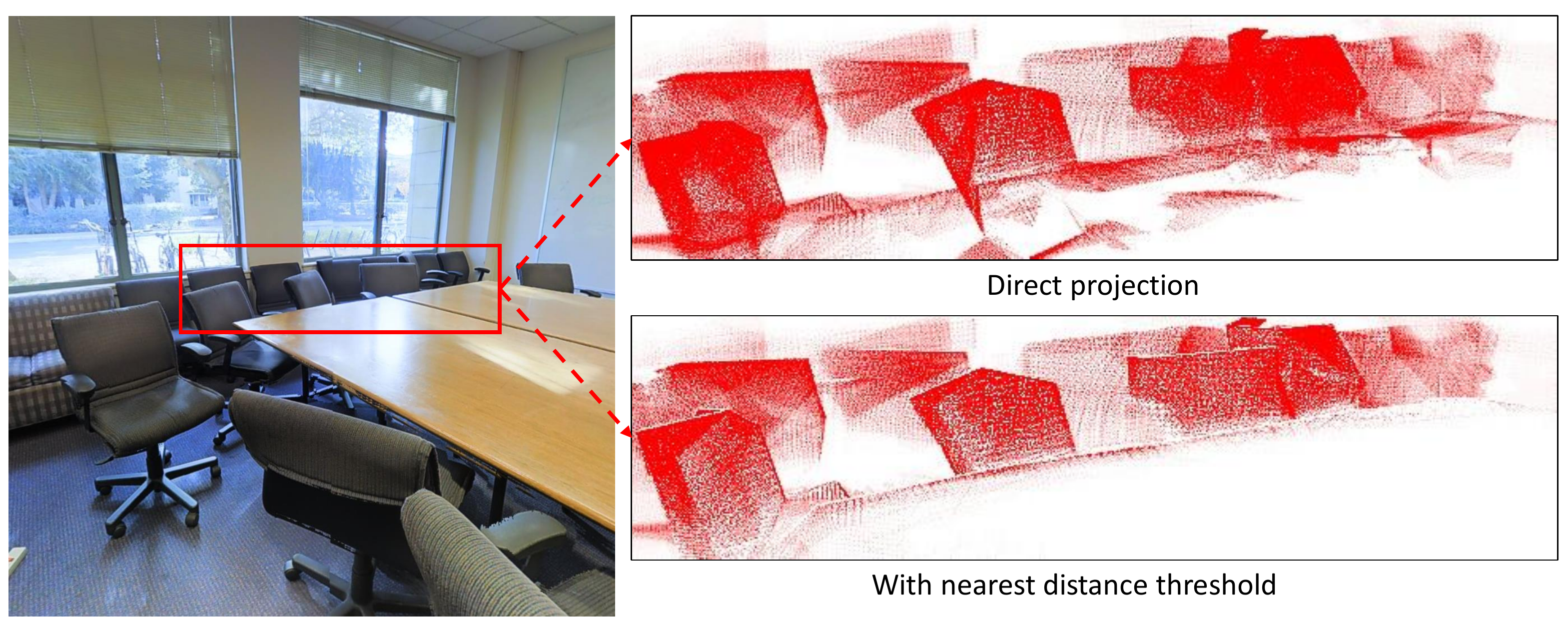}} 
   \end{center}
\caption{Illustration of the proposed nearest distance threshold. As shown in the left RGB image, some parts of the chairs are occluded under the table. With direct projection, the occluded parts of the chairs are still wrongly projected. After applying our nearest distance threshold, occluded areas are properly ignored. 
}
   \label{fig:occlusion}
\end{figure}

\textbf{Nearest Distance Threshold}
Occlusion may occur during the back projection, 
\ie, when objects are overlapped facing a camera, points belong to both visible objects and occluded objects are projected to the same region. 
To address this issue, we propose a nearest distance threshold.
by using the depth map. 
Depth represents the nearest surface to the camera, all 3D points behind the surface are occluded which should not be projected to 2D. 
Concretely, for a 3D point $P_j$,
we calculate its distance to the camera as $z_j$. 
Then we project the 3D point back to 2D at camera viewpoint $i$ at position $x_j^i$ and obtain the depth threshold $d_j^i$ at that position. Only points with $z_j <= d_j^i$ can be projected to generate objectness score masks.
Sample results are shown in Figure \ref{fig:occlusion}.

\textbf{Segment Proposals Refinement}
In this section, we propose a method to refine scattered objectness probability masks into segment proposals.
We take the chair class as an example and display results in Figure \ref{fig: proposal refine}.
As shown in the third column of Figure \ref{fig: proposal refine}, the objectness probability mask displays accurate object localization and objectness probability.
However, the projected masks are sparse and cannot directly be used as a supervision. 
To address this issue, we adopt a morphological operation followed by a fully-connected CRF \cite{krahenbuhl2011efficient} to recover dense segment proposals from the scattered masks. 

First, we follow \cite{otsu1979threshold} to binarize the projected objectness probability mask, then apply an image close operation. Then, we follow \cite{chen2014semantic} to adopt a fully connected CRF \cite{krahenbuhl2011efficient} to refine local boundary areas of our segment proposals. 
Referring to the fourth column of Figure \ref{fig: proposal refine}, it shows the recovery of image object boundaries based on 2D features, resulting in accurate segment proposals.

\begin{figure}
\setlength{\abovecaptionskip}{1mm}  
   \begin{center}
   {\includegraphics[width=0.9\linewidth]{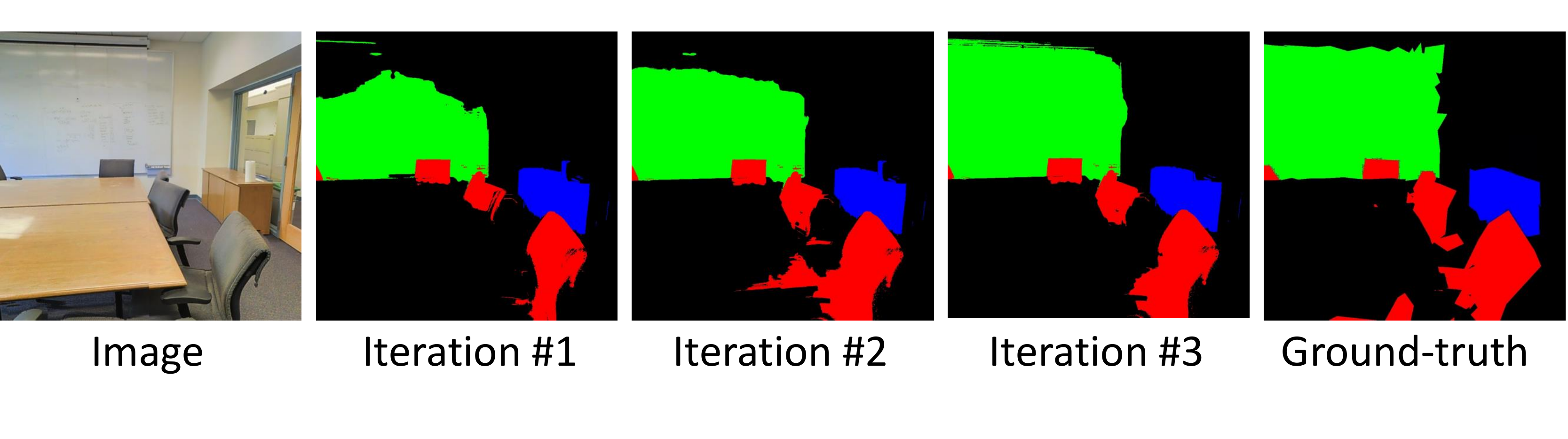} } 
   \end{center}
\caption{Recursive updating of
the segment proposals, which
are progressively refined in each iteration, and then treated as
supervision for the next iteration.}
   \label{fig: recursive}
\end{figure}

\subsection{Recursive Procedure}
\label{recursive procedure}
We observe that the generated segment proposals capture the object shape significantly better than bounding boxes, 
which inspires us to adopt a recursive training procedure of transductive segmentation.

First, we generate segment proposals with bounding box labels. Then, after fully training the segmentation network with the segment proposals, the segment predictions are fed back into our pipeline to generate new segment proposals, where the input is segmentation masks 
instead of bounding boxes.
New segment proposals are then used as supervision for the next iteration of training.
By cycles of segment proposal generation, learning a semantic segmentation network, and using the learned network to find improved masks on previously seen images, we iteratively improve segment proposals and segmentation network.

Moreover, we use the
bounding box labels to constrain our network predictions, \ie, bounding box labels ensure that regions outside do not contain any object. 
So we employ bounding box masks to apply binary masking on our network predictions, which effectively removes false positive areas:
$\Phi_p = X_b \otimes X_p$,
where $\otimes$ means spatial-wise masking, $X_b$ denotes masks generated by bounding boxes, $X_p$ denotes original network predictions. We feed the masked network predictions $\Phi_p$ into our segment proposal generation module as shown in the recursive connection in Figure \ref{fig:method_pipeline}.

\subsection{Data Annotation}
\label{Data annotation}
Currently, there exists no dataset available with multi-view camera parameters, depth data, and bounding box labels. Hence, we augment a subset of dataset 2D-3D-S \cite{2017arXiv170201105A} with hand labeled bounding boxes.
2D-3D-S \cite{2017arXiv170201105A} is an indoor dataset with multiple modalities from 2D, 2.5D and 3D domains, with instance-level semantic and geometric annotations. The dataset is collected in 6 large-scale indoor areas that originate from 3 buildings.
The dataset provides a corresponding depth map, camera parameters for each 2D image, we adopt them as 3D information. They also provide 2D segmentation ground-truth projected from semantically labeled 3D mesh model.
We pick four common indoor object classes (with clutter as the fifth class) to validate our method, including chair, bookcase, sofa and board as they are well defined and suitable for bounding box labels.

In this paper, we use data from area 1 to validate our method. Since there is no dataset contains both multi-view information and bounding box labels.
We manually label bounding boxes on a subset of 1822 images and obtain segment proposals on all training images by adopting our proposed algorithm. 
Here we introduce the process that how we select a subset of images.
First, since 2D images are separated by rooms, in order to obtain more object instances, we choose rooms where our objects are present and correctly labeled in the dataset.
Second, in each room, there are several camera locations, we ensure that the labeled images include images with views of the objects from different camera viewpoints to support our assumption of images over a wide baseline.
Then, the generated segment proposals are used as a supervision signal during training. It is worth noting that the annotation signal is entirely derived from the box-level labels. 
Moreover, our bounding box labels can be used to train a prior 2D detector, \ie, we can automate bounding boxes generation on more data, which further extends our method to a larger scale of data.

\section{Experiments}
\label{experiments}
\subsection{Setup} 
We evaluate the proposed method on the 2D-3D-S \cite{2017arXiv170201105A} dataset. The dataset and annotation details are introduced in Sec.~\ref{Data annotation}.
We adopt the publicly available DeepLabV3+ \cite{Chen_2018_ECCV} model as our backbone network. 
DeepLabv3+ is a recent state-of-the-art segmentation pipeline that uses a ResNet head \cite{He_2016_CVPR}, pre-trained on ImageNet \cite{deng2009imagenet}.
We keep network structure unchanged, and train models under different supervision conditions to validate the effectiveness of our method. 
During training, the initial learning rate is 0.01 and is decreased by a factor of 10 after every 10 epochs. SGD is used as our optimizer with momentum of 0.9 and weight decay of 0.0001. All the training data are augmented by random cropping and horizontal flipping. We do not adopt a fully-connected CRF for post-processing of our network predictions.
Results reported in Table \ref{table: baseline}, Table \ref{table: Performance of test set} and Figure \ref{fig: ratio} are all from the first iteration without our recursive method.
The evaluation performance is measured in terms of pixel intersection-over-union (mIoU).
All experiments were conducted using PyTorch.

\subsection{Comparison with Other Methods}
\label{baseline models}
In Table \ref{table: baseline}, comparisons are made to evaluate the impact of different levels of supervision.
In our own method, we label bounding boxes on 1822 images and use our approach to obtain segment proposals on 4028 images as our training set, the performances are evaluated on a randomly selected validation set without overlapping without the training set.
As a naive baseline, \emph{Bounding boxes 1822},
we fill the bounding boxes of the labeled images as masks and use the filled masks as supervision.
We form a second baseline, \emph{CRF refined boxes 1822}, by
directly applying a CRF on the bounding boxes to generate segment proposals. 
Moreover, 
we adopt GrabCut \cite{rother2004grabcut} to directly extract segment proposals from bounding boxes as \emph{GrabCut 1822}.
Then we evaluate performance in fully-supervised mode where the ground-truth is provided by the 2D-3D-S \cite{2017arXiv170201105A}. We report performances of the models supervised with both 1822 pixel-wise ground-truth, \emph{pixel-wise 1882} and full 4028 pixel-wise ground-truth, \emph{pixel-wise 4028}. 
We also report the performance in semi-supervised mode. We randomly select 400 images from the 4028 training set images, replace supervision with the ground-truth provided by \cite{2017arXiv170201105A}.
which is called \emph{semi 1822+400} in Table \ref{table: baseline}.

In addition, we compare with two state-of-the-art methods, \ie, SDI \cite{khoreva2017simple}, WSSL \cite{papandreou2015weakly}. 
For bounding box based weakly supervised semantic segmentation, the state-of-the-art methods are: \cite{song2019box}, BoxSup \cite{dai2015boxsup}, SDI \cite{khoreva2017simple} and WSSL \cite{papandreou2015weakly}. However, \cite{song2019box,dai2015boxsup,khoreva2017simple} did not release the official implementation code. Further, the released code for \cite{papandreou2015weakly} was based on another deep framework.
We re-implement SDI \cite{khoreva2017simple} and WSSL \cite{papandreou2015weakly} with Pytorch
as their performance is still competitive with SOTA and have clear and explicit implementation details. Our implementations will be made publicly available.

\begin{table}[!b]
\centering
\begin{tabular}{M{1.2cm} | c | M{1cm} | M{1cm} | M{1cm} | M{1.5cm} | M{1.2cm} | M{1cm} } 
 \hline
 \textbf{Modes} & \textbf{Method}  & \textbf{Sofa} & \textbf{Board} & \textbf{Chair} & \textbf{Bookcase} & \textbf{Clutter} &\textbf{mIoU} \\ 
 \hline
 \multirow{2}{*}{Full} & \multicolumn{1}{l}{Pixel-wise 4028} & \multicolumn{1}{l}{62.77} & \multicolumn{1}{l}{89.57}& \multicolumn{1}{l}{70.58}& \multicolumn{1}{l}{77.77}& \multicolumn{1}{l}{96.23} & \multicolumn{1}{l}{79.38} \\\cline{2-8}
                                 & \multicolumn{1}{l}{Pixel-wise 1822} & \multicolumn{1}{l}{54.44} & \multicolumn{1}{l}{85.29}& \multicolumn{1}{l}{66.78}& \multicolumn{1}{l}{74.64}& \multicolumn{1}{l}{95.37}& \multicolumn{1}{l}{75.30} \\\hline \hline

 \multirow{5}{*}{Box} & \multicolumn{1}{l}{Bounding boxes 1822}& \multicolumn{1}{l}{34.23} & \multicolumn{1}{l}{71.23}& \multicolumn{1}{l}{45.30}& \multicolumn{1}{l}{67.65}& \multicolumn{1}{l}{90.46} & \multicolumn{1}{l}{61.78} \\\cline{2-8}
                                 & \multicolumn{1}{l}{CRF refined boxes 1822} & \multicolumn{1}{l}{31.47} & \multicolumn{1}{l}{73.13}& \multicolumn{1}{l}{49.09}& \multicolumn{1}{l}{68.26}& \multicolumn{1}{l}{91.48}& \multicolumn{1}{l}{62.69} \\\cline{2-8}
                                  & \multicolumn{1}{l}{GrabCut 1822} & \multicolumn{1}{l}{41.38} & \multicolumn{1}{l}{79.79}& \multicolumn{1}{l}{56.91}& \multicolumn{1}{l}{70.76}& \multicolumn{1}{l}{93.89}& \multicolumn{1}{l}{68.55} \\\cline{2-8}
                                 & \multicolumn{1}{l}{WSSL \cite{papandreou2015weakly} 1822} & \multicolumn{1}{l}{55.52} & \multicolumn{1}{l}{75.58}& \multicolumn{1}{l}{53.43}& \multicolumn{1}{l}{67.14}& \multicolumn{1}{l}{93.95}& \multicolumn{1}{l}{69.06} \\\cline{2-8}
                                & \multicolumn{1}{l}{SDI \cite{khoreva2017simple} $M \cap G+$ 1822} & \multicolumn{1}{l}{47.28} & \multicolumn{1}{l}{81.10}& \multicolumn{1}{l}{56.47}& \multicolumn{1}{l}{66.94}& \multicolumn{1}{l}{93.74}& \multicolumn{1}{l}{69.11} \\\cline{2-8}
                                 & \multicolumn{1}{l}{\textbf{Ours 1822}} & \multicolumn{1}{l}{63.24} & \multicolumn{1}{l}{81.23}& \multicolumn{1}{l}{60.31}& \multicolumn{1}{l}{62.10}& \multicolumn{1}{l}{94.45}& \multicolumn{1}{l}{\textbf{72.27}} \\\hline \hline    
 \multirow{1}{*}{Semi} & \multicolumn{1}{l}{Semi 1822+400} & \multicolumn{1}{l}{61.98} & \multicolumn{1}{l}{80.33}& \multicolumn{1}{l}{62.77}& \multicolumn{1}{l}{67.09}& \multicolumn{1}{l}{94.64}& \multicolumn{1}{l}{73.36} \\\hline
\end{tabular}
\caption{Comparison of performances under different supervision conditions.
The number after the method name means the number of images with human annotations that were used in this setting. \emph{$M \cap G+$}: using the masks where both MCG and GrabCut agree.
In bold is the best performing of the bounding box supervised methods. Our method outperforms competing box supervised methods and is midway to the fully supervised methods.
}
\label{table: baseline}
\end{table}

Table \ref{table: baseline} shows the results of our segment proposals compared with other methods. 
The bounding-box based baseline achieves 61.78 of mIoU while CRF refined bounding boxes improve the performance to 62.69.
To explore the upper-bound on weakly supervised performance, we include results with a fully supervised model, trained using
1822 and 4028 pixel-wise ground-truth images respectively, the scores are 75.30 and 79.38.
Our method achieves 72.27, which outperforms all the compared bounding-box based methods with a clear margin and is approaching the 1822 pixel-wise supervised baseline.
This validates that our proposed method is effective. By labeling a subset of the dataset, we can extract accurate segment proposals on labeled and unlabeled images. 
Finally, in semi-supervised mode, we achieve 73.36 mIoU through replacing segment proposals of 400 images (10\% of the training set) with pixel-level ground truth, where the performance is comparable with fully supervised models.
The performance of our semi-supervised model indicates that we can achieve even better performance if additional supervision information is provided.

\begin{figure}[!t]
\setlength{\abovecaptionskip}{1mm}  
   \begin{center}
   {\includegraphics[width=0.90\linewidth]{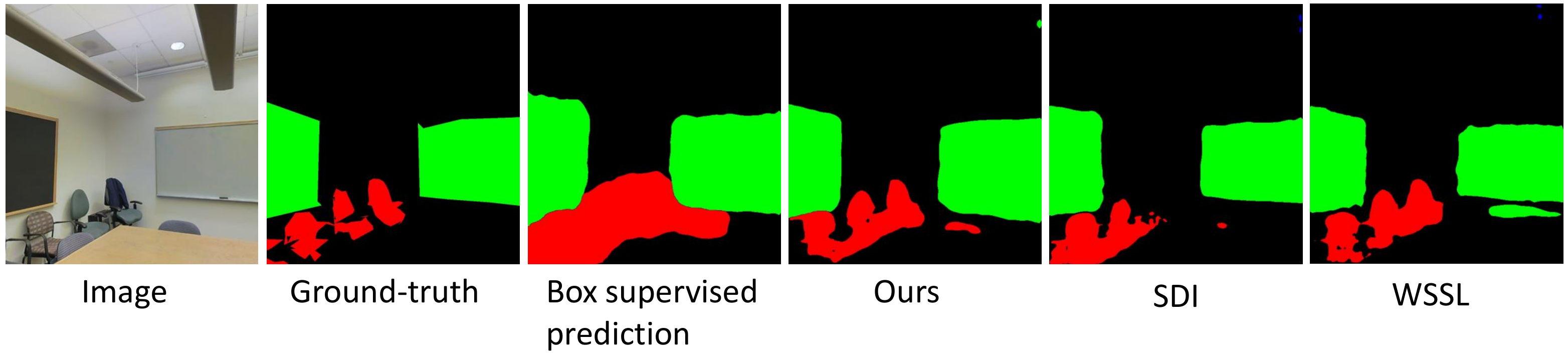}} 
   \end{center}
\caption{Examples of semantic segmentation prediction results. 
}
   \label{fig: prediction}
\end{figure}

\begin{table}[!t]
\centering
\begin{tabular}{c | c | c} 
\hline
 \textbf{Modes} & \textbf{Supervision} &\textbf{mIoU} \\ 
 \hline
 \multirow{1}{*}{Full} & \multicolumn{1}{l}{Pixel-wise 1822} & \multicolumn{1}{l}{63.62}  \\\hline \hline
                                          
 \multirow{3}{*}{Box} & \multicolumn{1}{l}{Bounding boxes 1822} & \multicolumn{1}{l}{53.57} \\\cline{2-3}
                                 & \multicolumn{1}{l}{CRF refined boxes 1822} & \multicolumn{1}{l}{54.82} \\\cline{2-3}
                                 & \multicolumn{1}{l}{\textbf{Ours 1822}} & \multicolumn{1}{l}{\textbf{57.61}} \\\hline \hline   
                                 
 \multirow{1}{*}{Semi} & \multicolumn{1}{l}{Semi 1822+1000} & \multicolumn{1}{l}{59.24} \\\hline
\end{tabular}
\caption{Comparison of performances on the test set.
}
\label{table: Performance of test set}
\end{table}

\subsection{Experiments on Data from Unseen Areas}
\label{Experiments on Data from Other Areas}
In our procedure, we label a subset of all images and get segment proposals for all those images by adopting 3D information. However, images from the same room may view the same object instances across the training and validation sets.
To validate the effectiveness of our proposed method, we randomly select 2000 images from new areas to assemble a test set. These new areas are \enquote{unseen}, which means none of the images in these areas are labeled nor seen in the training or validation. We test our trained model on this test set without fine-tuning.
As shown in Table \ref{table: Performance of test set}, our method outperforms the model trained with only bounding box masks, and achieves comparable performance to the fully supervised model.
Thus, it validates that our method captures precise object information, and provides similar general class-wise features as the pixel-wise ground-truth.

\subsection{Recursive Training Performance}
\begin{table}[!t]
\centering
\begin{tabular}{c | c | c | c} 
 \hline
 \textbf{Supervision} & \textbf{Iteration 1} & \textbf{Iteration 2} & \textbf{Iteration 3} \\ 
 \hline
 Ours & 57.61  & 59.12 & 59.83
 \\
 \hline
\end{tabular}
\caption{Evaluate effectiveness of the recursive training method.}
\label{table: recursive}
\end{table}

\label{recursive training performance}
As introduced in Sec. \ref{recursive procedure}, we may apply our proposed method in a recursive manner. In the recursive procedure, the network predictions are fed into our segment proposal generation pipeline
to generate new segment proposals.
Although we achieve good results in the first iteration, the recursive training process transductively refines the new segment proposals and improves our network progressively.
As shown in Table \ref{table: recursive}, we evaluate performance on the test set, the performance in every iteration is gradually improved. 

\begin{figure}[t!]
\setlength{\abovecaptionskip}{1mm}  
  \begin{center}
  {\includegraphics[width=0.54\linewidth]{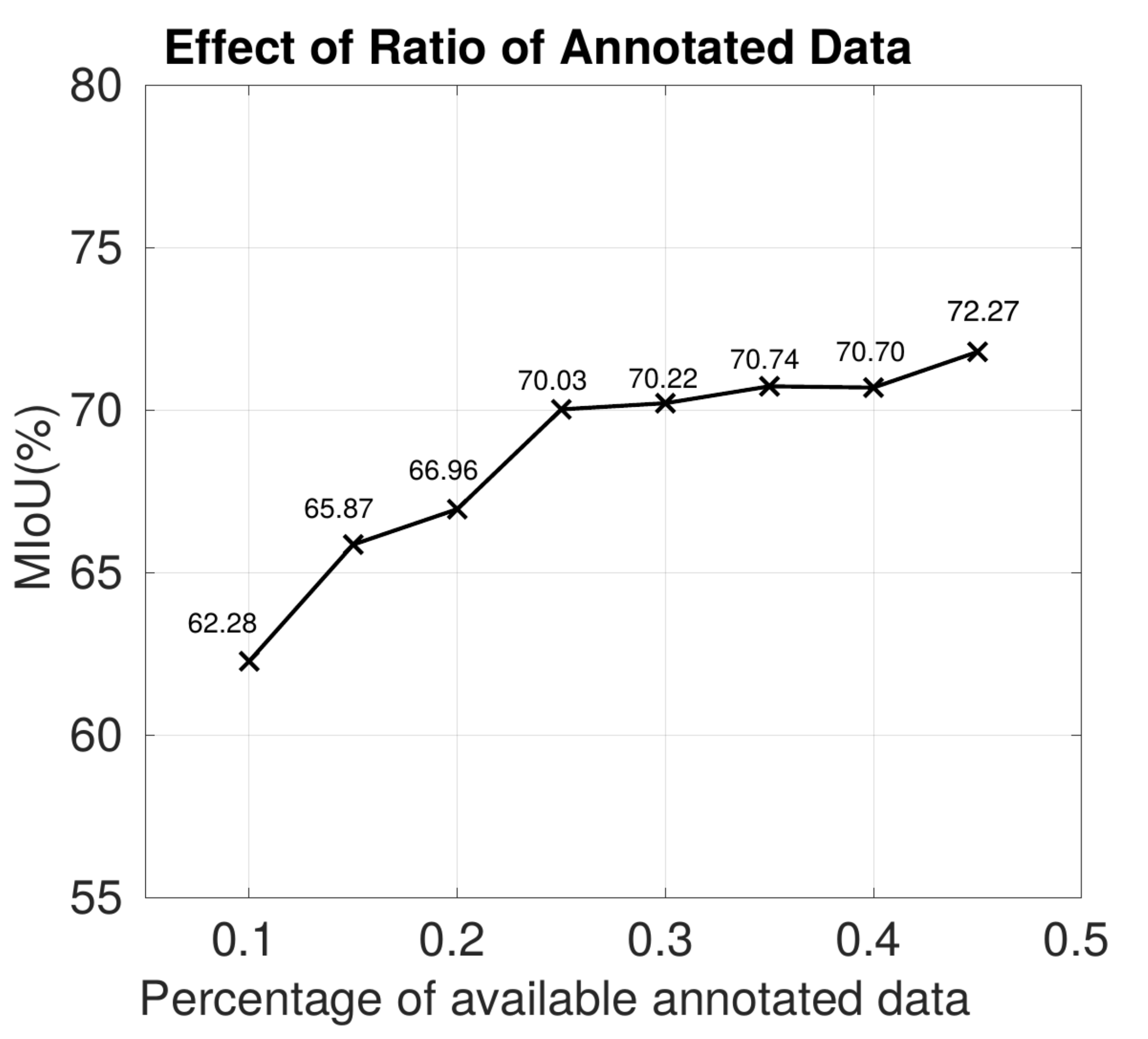} } 
  \end{center}
\caption{Performance of our method when we randomly select different percentages of available annotated data. Reasonable segmentation is possible when just 10 percent of the training set is annotated. We can observe clear improvement when more annotation is provided, trailing off above 25 percent.}
  \label{fig: ratio}
\end{figure}

\subsection{Ablation: Effect of Ratio of Annotated Data}
\label{section: effect of ratio}
Our proposed method labels only a small portion of the dataset and obtains segment proposals on all images, which drastically decreases the cost of annotation.
In this section, we investigate how the percentage of available annotated data affects performance, and attempt to achieve a balance between annotation cost and network performance. 1822 labeled images make up
45\% of 4028 training images. 
We randomly select images from these
to get different ratios of available labeled images. 
Selected labels are then fed
into the same pipeline to generate segment proposals on the training set and report results in Figure \ref{fig: ratio}.
As shown,
more annotation data leads to manifest performance improvements which indicates the effectiveness of our method. 
When trained using only 25\% of training images, 
we achieve 70.03 mIoU which already
outperforms all competing methods (trained on 45\%).
Moreover, our method still gets a reasonably good results 
when only 10\% of the training data is labeled with bounding boxes. 

\section{Conclusion}
\label{conclusion}
In this paper, we propose a novel 3D weakly supervised semantic segmentation approach, which incorporates box-level labels with corresponding 3D information.
By only labeling a small number of images with bounding boxes, our approach extracts segment proposals on labeled and unlabeled images. Then we use the obtained segment proposals to train a semantic segmentation model.
Moreover, our method can work in a recursive manner,
which further refines our segment proposals. 
Our proposed method achieves competitive semantic segmentation results with less annotation effort.
We evaluate the proposed method on the 2D-3D-S \cite{2017arXiv170201105A} dataset, extensive experimental results show that our proposed method is effective.
Our annotations and source code will be made publicly available for future research endeavors in this field. 

\bibliographystyle{splncs}
\bibliography{egbib}

\end{document}